\documentclass[conference]{IEEEtran}
\IEEEoverridecommandlockouts
\usepackage{cite}
\usepackage{amsmath,amssymb,amsfonts}
\usepackage{algorithmic}
\usepackage{graphicx}
\usepackage{textcomp}
\usepackage{xcolor}
\usepackage{caption}
\usepackage{subcaption}
\usepackage{url}
\usepackage{hyperref}
\def\BibTeX{{\rm B\kern-.05em{\sc i\kern-.025em b}\kern-.08em
    T\kern-.1667em\lower.7ex\hbox{E}\kern-.125emX}}
    
\begin{document}

\title{Trajectory-based Algorithm Selection with Warm-starting
\thanks{Anja Jankovic (Email: Anja.Jankovic@lip6.fr) 
and Carola Doerr (Email: Carola.Doerr@lip6.fr) 
are with Sorbonne Université, CNRS, LIP6, Paris, France. 
Diederick Vermetten (Email: d.l.vermetten@liacs.leidenuniv.nl) 
and Jacob de Nobel (Email: j.p.de.nobel@liacs.leidenuniv.nl) 
are with Leiden University, The Netherlands. 
Ana Kostovska (Email: ana.kostovska@ijs.si) 
is with the Department of Knowledge Technologies, Jo\v{z}ef Stefan Institute, Ljubljana, Slovenia as well as with the Jo\v{z}ef Stefan International Postgraduate School, Ljubljana, Slovenia. 
Tome Eftimov (Email: tome.eftimov@ijs.si) 
is with the Computer Systems Department, Jo\v{z}ef Stefan Institute, Ljubljana, Slovenia.}
\thanks{Our work is supported by the Paris Ile-de-France region, by projects from the Slovenian Research Agency (research core funding No. P2-0098, P2-0103, the young researcher grant No. PR-09773 to AK, and researcher funding No. N2-0239), as well as by the EC through grant No. 952215 (TAILOR)}} 
% \thanks{}
% }
% \author{\IEEEauthorblockN{Anonymous Authors}}
\author{\IEEEauthorblockN{Anja Jankovic, 
        Diederick Vermetten,
        Ana Kostovska,
        Jacob de Nobel, 
        Tome Eftimov
        and Carola Doerr}
}

\maketitle

\begin{abstract}
Landscape-aware algorithm selection approaches have so far mostly been relying on landscape feature extraction as a preprocessing step, independent of the execution of optimization algorithms in the portfolio. This introduces a significant overhead in computational cost for many practical applications, as features are extracted and computed via sampling and evaluating the problem instance at hand, similarly to what the optimization algorithm would perform anyway within its search trajectory. As suggested in~[Jankovic et al., EvoAPP 2021], trajectory-based algorithm selection circumvents the problem of costly feature extraction by computing landscape features from points that  a solver sampled and evaluated during the optimization process. Features computed in this manner are used to train algorithm performance regression models, upon which a per-run algorithm selector is then built.

In this work, we apply the trajectory-based approach onto a portfolio of five algorithms. We study the quality and accuracy of performance regression and algorithm selection models in the scenario of predicting different algorithm performances after a fixed budget of function evaluations. We rely on landscape features of the problem instance computed using one portion of the aforementioned budget of the same function evaluations. Moreover, we consider the possibility of switching between the solvers once, which requires them to be warm-started, i.e. when we switch, the second solver continues the optimization process already being initialized appropriately by making use of the information collected by the first solver. In this new context, we show promising performance of the trajectory-based per-run algorithm selection with warm-starting.
\end{abstract}


\section{Introduction}
\label{sec:intro}

Optimization is a central notion across a broad range of scientific disciplines and real-world applications. Finding an optimal solution for a given problem is often not a straightforward process, as problems are typically computationally hard or otherwise intractable. In many concrete use cases, knowledge about the inherent nature of the problem is very limited, which renders formal problem modeling impossible. Under these circumstances, users are required to treat such problems as a black box. \emph{Black-box optimization} (BBO) provides techniques that are able to generate good solutions for these problems in a reasonable time. These techniques, known as BBO algorithms, operate via an iterative process of sampling and evaluating solution candidates and using the knowledge obtained in the previous iterations to guide the search towards more promising alternatives, until eventually converging to the best estimate of an optimal solution.

Due to the plethora of existing optimization problems, different BBO algorithms have been developed to this day. Various underlying operating mechanisms of these algorithms yield their complementary behavior on different problems. This large algorithmic variety poses a meta-optimization problem in achieving peak performance~\cite{Rice76}: how does one select the most efficient algorithm for a given problem instance? 

In recent years, significant research focus has been put on algorithm selection approaches that make use of the knowledge about the problem instance landscape to base the decision about which algorithm to use in that particular situation~\cite{KerschkeT19, MunozASSurvey, KerschkeHNT19survey}. \emph{Landscape-aware algorithm selection} generally relies on an important preprocessing step of extracting information from the problem instance landscape (independently of the optimization process). A huge challenge in this regard is the overhead computational cost induced by this preprocessing step, as further resources are spent on sampling and evaluating search points to first characterize the landscape, but are not at all considered while executing the algorithm on a problem instance. In many practical applications, users cannot afford to spend those additional function evaluations prior to optimizing, as they can be very expensive (e.g., crash tests or clinical studies). The approach suggested in~\cite{DBLP:conf/evoW/JankovicED21} offers a convenient alternative perspective in which the information about the problem instance landscape is extracted via samples and their function evaluations performed anyway by the algorithm. This framework shows preliminary potential in circumventing the preprocessing step altogether, and might present a step forward in the direction of designing an efficient fully dynamic algorithm selection model. However, an important open question from~\cite{DBLP:conf/evoW/JankovicED21} remains: how do we make use of the trajectory-based information of a default algorithm to predict performances of other algorithms? We tackle it with this work.

In this paper, we extend the trajectory-based landscape-aware approach to a portfolio of five widely used black-box optimization algorithms, and we consider that we can switch between the algorithms once during the optimization process. Put differently, we start optimizing by the base algorithm (\emph{A1}) while collecting the landscape data from the algorithm trajectory, then switch to another algorithm (\emph{A2}) that is \emph{warm-started}, i.e. appropriately initialized by making use out of the gathered information from \emph{A1}. We first show that we are able to get decent predictions of the \emph{A2} performance using \emph{A1}'s trajectory-based landscape characterization, upon which we then build an algorithm selection model. At higher levels of performance complementarity between the algorithms (which coincide with larger total budgets for running the algorithms), the algorithm selector outperforms any of the standalone algorithms in terms of loss against the true best algorithm for each particular problem instance. We finally highlight some advantages as well as drawbacks with respect to different parts of the adopted pipeline and present some important challenges for future work.

\textbf{Outline of the Paper:}
% The paper is organized as follows: t
Throughout Sec.~\ref{sec:background}, we position the work in the context of numerical black-box optimization, introduce the benchmark collection (\ref{ssec:BBOB}), the algorithm portfolio selected for our work (\ref{ssec:portfolio}), and the notion of warm-starting (\ref{ssec:warmstarting}), recap the state-of-the-art in per-instance algorithm selection (\ref{ssec:AS}) and establish the ground for the trajectory-based approach in analyzing problem instance landscapes (\ref{ssec:ELA}). We give an overview of the full experimental setup in Sec.~\ref{sec:setup}, focusing on each component of the trajectory-based algorithm selection framework applied to a diverse algorithm portfolio. We present and discuss the main findings of our study in Sec.~\ref{sec:results} and provide critical assessment of this approach in Sec.~\ref{sec:limitations}. Finally, in Sec.~\ref{sec:conclusions}, we wrap up with %some promising avenues and possibilities for addressing 
open questions that merit further attention.

\textbf{Reproducibility and Additional Figures:} 
To ensure that the work shown in this paper is reproducible~\cite{ManuelReproducibilityTELO}, all data and code used is made available on \texttt{figshare}~\cite{figshare_figures_noano}.
This includes figure generation code for figures which have not been included here because of the limited space available. For ease of viewing, these additional figures can also be viewed in this same repository. In particular, the losses of our model for different budgets can be found there.

%%%%%%%%%%%%%%%%%%%%%%%%%%
%%%%%%%% BACKGROUND %%%%%%
%%%%%%%%%%%%%%%%%%%%%%%%%%
\section{Background and Selected Benchmark Environment}
\label{sec:background}

Black-box optimization algorithms are a wide-ranging family of adaptive sampling-based strategies.
% Once initialized, they generally perform an iterative search procedure until reaching a predetermined stopping criterion.
Many of these algorithms can adjust their behavior during the optimization process by taking into account the information gathered during the run. This is especially useful as typically different phases of the optimization process require different search behavior (conveniently illustrated in the trade-off between exploration and exploitation). Moreover, different algorithms employ different search routines which can be more or less beneficial depending on a given scenario. A large open question that stems from this observation is to detect which algorithm is the most suitable for a given stage of the optimization process. First steps into this direction have shown potential in switching the algorithm once during the run itself~\cite{VermettenRBD19}. However, this potential remains largely untapped if we restrain ourselves to a family of similar algorithms~\cite{modCMA}. Having a wider set of different solvers might help bypass this obstacle.

\subsection{The BBOB Problem Collection}
\label{ssec:BBOB}
In the context of numerical black-box optimization, assessing the performance of different algorithms on different problem instances is largely facilitated by the existence of well-established \emph{benchmark problem collections}. In this study, we rely on one such collection, the BBOB noiseless testbed from the COCO environment~\cite{cocoplat}. The BBOB collection comprises 24 functions. For each of these functions, multiple problem instances are available. 
We conveniently generate, access and analyze the benchmark data via the \texttt{IOHprofiler} platform~\cite{IOHprofiler20}.  

\subsection{Algorithm Portfolio}
\label{ssec:portfolio}
% With respect to the choice of the algorithm portfolio, f
Following suggestions from~\cite{DBLP:conf/gecco/VermettenWBD20} and related works, we opt for five algorithms that are frequently used to tackle numerical black-box problems: 
\begin{itemize}
    \item BFGS (Broyden-Fletcher-Goldfarb-Shanno~\cite{bfgs1,bfgs2,bfgs3,bfgs4}): a Quasi-Newton method that approximates the Jacobian or the Hessian instead of actually computing them.
    \item CMA-ES (Covariance Matrix Adaption - Evolution Strategy~\cite{cmaes}): a stochastic derivative-free numerical optimization algorithm that iteratively samples the population from a multivariate normal distribution and updates the shape of the said distribution with the information gathered while running.
    \item DE (Differential Evolution~\cite{DE}): a population-based algorithm that samples candidates purely based on numerical differences between existing population members.
    \item MLSL (Multi-Level Single Linkage~\cite{mlsl1,mlsl2}): an %systematic trajectory 
    algorithm that combines global search phases (based on clustering) with more focused, local search procedures.
    \item PSO (Particle Swarm Optimization~\cite{pso}): simulates particles moving around the search space based on their individual velocity, determining both speed and direction, motivated by the swarm behavior of some animal species.
\end{itemize}

\subsection{Warm-starting}
\label{ssec:warmstarting}

To enable switching from CMA-ES to each of the five algorithms mentioned previously, we make use of some basic warm-starting strategies. The most intuitive version of warm-starting is to inherit the best-so-far solution and use this as the starting point for the second algorithm. This point can then be used as a center to initialize the new population around, as is done by CMA-ES, DE and PSO. For switching to BFGS, we can use the covariance matrix directly, by using this as the inverse of the Hessian~\cite{DBLP:journals/tcs/ShirY20}. We should note that by warm-starting the CMA-ES using the same procedure as the other population-based approaches, we lose a significant amount of information, which might lead to worse performance than expected if we had not switched. However, it does mean that the approach is easier to extend, since we can modify the starting algorithm while having minimal impact on the procedure as a whole. 

\subsection{Per-Instance Algorithm Selection}
\label{ssec:AS}

As mentioned in Sec.~\ref{sec:intro}, given an optimization problem, its specific instance that needs to be solved, and a set of algorithms that can be used to solve it, the so-called
\emph{per-instance algorithm selection (PIAS)} problem arises: how to determine which of those algorithms can be expected to perform best on that particular instance? In other words, one is not interested in having an algorithm recommendation for a whole problem class (such as TSP or SAT in the discrete domain), but for a specific instance of some problem. A large body of work exists in this line of research~\cite{BischlMTP12, CossonDLATZ21, HutterKV19, KerschkeHNT19survey, LindauerHHS15, XuHHL12}, and they mostly rely on extracting the information about the problem instances beforehand, contrary to this study.

To assess the algorithm selector's quality, two standard baselines are used. The performance of a (hypothetical) perfect per-instance algorithm selector,
also known as the \emph{virtual best solver} (VBS) or the oracle selector, provides a lower bound
on the performance of any realistically achievable algorithm selector. The VBS always selects the true best algorithm per each problem instance. On the other hand,
a natural upper bound on the algorithm selector performance is provided by the \emph{single
best solver} (SBS), which is the algorithm with the best overall performance among all other
algorithms in the considered portfolio.
The ratio between VBS and SBS performances, also referred to as the \emph{VBS-SBS gap},
gives an indication of the performance gains that can be obtained by per-instance
algorithm selection in the best case. Consequently, the fraction of this gap closed by a
certain algorithm selector provides a measure of its quality~\cite{lindauer17}.

\subsection{Trajectory-Based Exploratory Landscape Analysis}
\label{ssec:ELA}

In order to represent the considered optimization problem instances in a suitable and useful way
for the algorithm selection pipeline, we shall want to quantify their different characteristics via appropriate measures. This is typically done by means of \emph{exploratory
landscape analysis} (ELA)~\cite{mersmann_exploratory_2011}. Problem instances are characterized by automatically computed ELA \emph{features} using information extracted via sampling and evaluating the problem. A vector of numerical ELA feature values is assigned to each instance and can be then used to train a predictive model that maps it to different algorithms' performances on the said instance. The feature extraction step is commonly considered to be independent from the optimization process, and diverse sampling strategies can be employed, %when gathering search points for feature computation, %e.g., uniform random sampling, Latin hypercube designs, as well as Sobol' sequences, 
see~\cite{renau2020exploratory} for a discussion. Conveniently, feature computation is done via the R package \texttt{flacco}~\cite{flacco}, and the ELA features we consider here are suggested in~\cite{DBLP:conf/evoW/JankovicED21}. % and further references.

However, as the nature of the knowledge needed to extract features and to optimize a problem instance is the same, the motivation arises to save computational resources from the preprocessing step by incorporating the feature extraction within the optimization process. The core idea is to utilize samples already evaluated by the algorithm to compute the landscape features (as seen locally on the algorithm's search trajectory)~\cite{DBLP:conf/evoW/JankovicED21}. We adopt this perspective in this paper, extending it to a more diverse portfolio. We collect the points sampled by the base algorithm and use so-computed features to predict the performance of another (warm-started) algorithm which continues the optimization process.  
% Preliminary results in~\cite{DBLP:conf/evoW/JankovicED21} suggest that the trajectory-based approach can lead to robust predictions and good quality algorithm selection built atop.

%The trajectory-based algorithm selection can hence be considered an extension of the per-instance algorithm selection towards a \emph{per-run algorithm selection}.

%%%%%%%%%%%%%%%%%%%%%
%%%%%%%% SETUP %%%%%%
%%%%%%%%%%%%%%%%%%%%%
\section{Experimental Setup}
\label{sec:setup}

As discussed in Sec.~\ref{ssec:portfolio}, in this work we make use of a small portfolio of algorithms. The specific portfolio is chosen based on the observed differences in their potential performance as either the first (\emph{A1}) or second (\emph{A2}) part of a dynamically switching algorithm~\cite{DBLP:conf/gecco/VermettenWBD20}. In particular, we consider the following algorithm implementations:
\begin{itemize}
    \item CMA-ES: %For our implementation of the CMA-ES, 
    we use the modular CMA-ES (\texttt{modCMA})~\cite{van_rijn_evolving_2016,nobel_modcma_assessing}, which implements a wide range of variants into one modular framework with default settings and  saturation as the boundary correction method.
    % . For this study, we use the default settings with saturation as the boundary correction method.
    \item DE: we use the \texttt{scipy}~\cite{2020SciPy-NMeth} implementation.
    \item PSO: we implement a basic version with clipped velocity to avoid exploding trajectories.
    \item MLSL: we implement the version described in~\cite{pal2013benchmarking}, using \texttt{scipy}'s version of BFGS for the local search procedure.
    \item BFGS: we adapt \texttt{scipy}'s implementation.
\end{itemize}

We then use the first five instances of each of the 24 BBOB functions mentioned in Sec.~\ref{ssec:BBOB} for our experiments.

Our dynamic algorithm follows a two-stage process; first it starts with the modular CMA-ES for $30\cdot D$ evaluations, rounded up to the nearest multiple of the used population size. This equates to $154$ evaluations for the $5$-dimensional version of the functions. 
After this point, the run is interrupted and the second algorithm is warm-started as described in Sec.~\ref{ssec:warmstarting}. This experiment is repeated 10 times on each of the first five instances of all 24 BBOB functions and for each of the five algorithms, resulting in a total of $6\,000$ runs. To execute this data collection, we used \texttt{IOHprofiler}~\cite{IOHprofiler}, which enabled us to keep track of the full search history, as well as performance data and the state variables, some of which are needed to warm-start the algorithms that we switch to. As our performance measure, we take the function value reached after a fixed number of function evaluations (i.e., the fixed-budget target precision). 

The study presented in~\cite{DBLP:conf/evoW/JankovicED21} also experimented with the use of state variables as features for the performance predictions. Since no significant advantage was observed in~\cite{DBLP:conf/evoW/JankovicED21} for these variables, we do not make use of them here in this work (apart from extracting the information that is needed to warm-start the algorithms after the switch). In order to allow for a comparison with such an approach, we have nevertheless recorded the state variables listed in~\cite{DBLP:conf/evoW/JankovicED21}; they can be found in the data record made available at~\cite{figshare_figures_noano}.

\subsection{Performance Data}
\label{ssec:raw}

\begin{figure*}
    \centering
    \includegraphics[width=\textwidth]{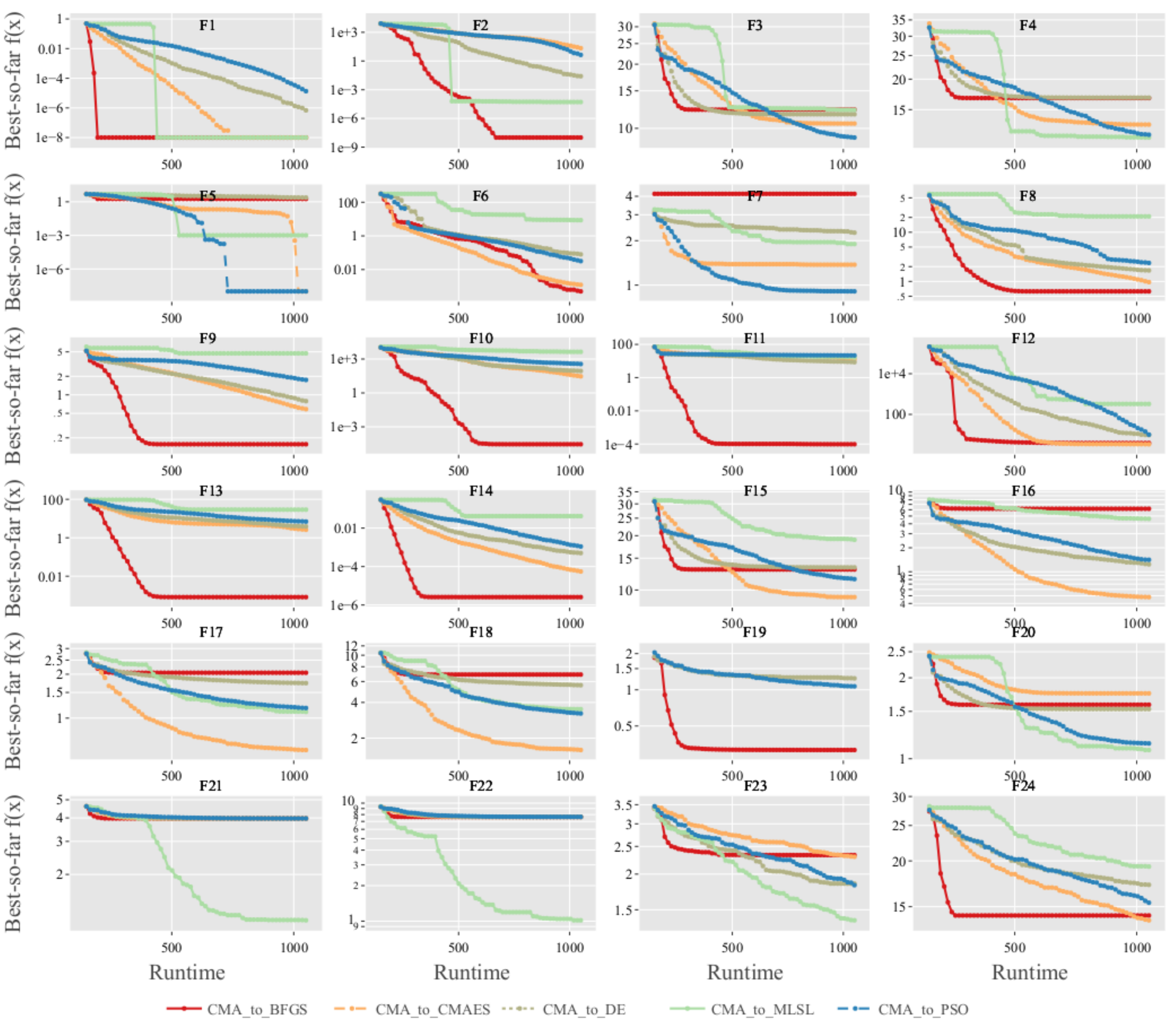}
    \caption{Mean best-so-far function value (target precision) reached by each of the five switching algorithms on all 24 BBOB functions. Each line corresponds to 50 runs: 10 on each of the first five instances of the function. Note that the first $154$ evaluations are identical for each algorithm, and are thus excluded from the figure. Figure generated using \texttt{IOHanalyzer}~\cite{IOHanalyzer}.}
    \label{fig:perf_curves}
\end{figure*}

For each switching algorithm, we collect a total of $1\,200$ runs. Since these runs all start with the same $154$ evaluations from the CMA-ES, we are mostly interested in the complementarity of their performance after this point. To visualize this, we show the evolution of the mean function value over time in Fig.~\ref{fig:perf_curves}. Here, we can see that for most of the unimodal functions, the switch to BFGS significantly outperforms all others, as would be expected, since the BFGS has the most involved warm-starting procedure. For the more complex functions, however, this initial benefit from switching to BFGS disappears after a while, with the other algorithm catching up and steadily overtaking it. This highlights a key aspect of the prediction problem, namely the allocation of budget to the second algorithm. The optimal switching algorithm for a total budget of $350$ can differ widely from one where the overall budget is $1\,050$ evaluations. Luckily, we can simulate the procedure for short budgets by cutting of the run at the required point and measuring the performance at that point, allowing us to investigate the impact of this overall budget is more detail. In particular, we consider the following set of \emph{A2} budgets for the second part of the search: $\{100,200,300,500,700,900\}$ function evaluations, while the \emph{A1} trajectory budget allocated for the feature extraction is fixed at $150$ function evaluations.

% \begin{figure}
%     \centering
%     \includegraphics[width=0.48\textwidth,trim=160mm 5mm 40mm 5mm,clip]{figures/Best_Alg_Matrix.pdf}
%     \caption{Heatmap showing in how many (out of $1\,200$) runs each algorithm is the best one to switch to, based on the amount of budget given to this second part of the search. Results are capped at $10^{-8}$ target precision, which can lead to ties. The number of best algorithms per budget can therefore be larger than $1\,200$.}
%     \label{fig:best_alg}
% \end{figure}

\subsection{Selection of Regression Models}
\label{ssec:regression}

% Here we describe the process of training and selecting the regression models for the algorithm performance prediction. 

We note here that, as we operate within a fixed-budget setting, target precision values rapidly get smaller as we converge to the optimum. Therefore, not only do we perform a classical regression on the actual data, but we also take into account the possibility of more accurately predicting these very small target precision values via training the regression model on the log-values of the same data. We then learn a separate regression model per each combination of the algorithm, considered budget and type of the target value (actual and log). This leaves us with a total number of 60 different regression models (5 algorithms $\times$ 6 budgets  $\times$ 2 targets). To learn the regression models, as classically suggested in the literature, we use the Random Forest (RF) algorithm as implemented in the Python package \texttt{scikit-learn}~\cite{pedregosa2011scikit} and perform hyperparameter tuning using the grid search methodology. We tune five different RF hyperparameters: (1) \emph{n\_estimators} -- the number of trees in the random forest; (2) \emph{max\_features} -- the number of features used for making the best split ; (3) \emph{max\_depth} -- the maximum depth of the trees; (4) \emph{min\_samples\_split} -- the minimum number of samples required for splitting an internal node in the tree; and (5) \emph{criterion} -- the function that measures the quality of a given split. The full list of tuned hyperparameters and their corresponding search spaces is given in Tab.~\ref{tab:hyperparameters}. 

To evaluate the predictive performance of the regression models, we employ the \emph{leave-one-group-out} strategy. Here, the groups are defined on the ID of the problem instance (1--5), which means we work with five different groups. We thus perform five iterations over the data, and we hold one instance out each time (all 10 runs included), train the model on the remaining data, and test on the test (hold-out) data. We use the $R^2$ score as an evaluation measure of predictive power of the models. Finally, to obtain the test error, we compute the average $R^2$ score over the five hold-out groups. The average $R^2$ scores for the regression models with actual target precision are given in Tab.~\ref{tab:score_unlog}, while  $R^2$ scores for the regression models with the log-target precision can be found in Tab.~\ref{tab:score_log}.

\begin{table}
\begin{center}
\caption{RF hyperparameter names and their corresponding values considered in the grid search.}
\label{tab:hyperparameters}
\begin{tabular}{ cc }
\hline
 Hyperparameter & Search space \\ 
\hline
 n\_estimators & $[100, 500, 1000]$\\
 max\_features & $[\textsc{auto}, \textsc{sqrt}, \textsc{log2}]$ \\ 
 max\_depth & $[4,8,15, \textsc{None}]$ \\
 min\_samples\_split & $[2, 5, 10]$ \\
 %criterion & $['squared\_error', 'absolute\_error', 'poisson']$ \\
 criterion & \begin{tabular}[x]{@{}c@{}}$[\textsc{squared\_error}, \textsc{absolute\_error},$\\$ \textsc{poisson}]$\end{tabular} \\
 \hline
\end{tabular}
\end{center}

\end{table}
\begin{table}
\begin{center}
\caption{$R^2$ scores for the regression models trained on the actual target precision for all considered \emph{A2} budgets.}
\label{tab:score_unlog}
\begin{tabular}{ ccccccc }
\hline
 Algorithm & 100 & 200 & 300 & 500 & 700 & 900 \\ 
\hline
BFGS &	0.0637 & 0.3059 & 0.4764 & 0.4854 &	0.4869 &	0.4860 \\
CMAES &	0.5030 & 0.1473 & 0.0993 & 0.2514 & 0.2353 & 0.1152 \\
DE & 0.1700	& 0.2699 & 0.1571 & 0.1322 & 0.0333 & -0.0321 \\
MLSL &	0.2410 & 0.2066 & 0.3142 & -0.1059 & -0.0641 & -0.0279 \\
PSO	& 0.5361 & 0.5694 & 0.5884 &	0.3919 & 0.1398 & -0.8812\\
 \hline
\end{tabular}
\end{center}
\end{table}

\begin{table}
\begin{center}
\caption{$R^2$ scores for the regression models trained on the log-target precision for all considered \emph{A2} budgets.}
\label{tab:score_log}
\begin{tabular}{ ccccccc }
\hline
 Algorithm & 100 & 200 & 300 & 500 & 700 & 900 \\ 
\hline
BFGS & 0.7016 & 0.6691 & 0.7073 & 0.7425 & 0.7492 & 0.7570 \\
CMAES & 0.6708 & 0.7006 & 0.7695 & 0.8423 & 0.8053 & 0.7894 \\
DE & 0.6721 & 0.6549 & 0.6324 & 0.6109 & 0.6194 & 0.6669 \\
MLSL & 0.7296 & 0.7277 & 0.8722 & 0.8687 & 0.8678 & 0.8688 \\
PSO	& 0.7205 & 0.7017 & 0.7980 & 0.9128	& 0.9137 & 0.8745 \\
\hline
\end{tabular}
\end{center}
\end{table}
We observe from Tab.~\ref{tab:score_unlog} and Tab.~\ref{tab:score_log} that the regression models for log-target precision generally outperform the models with the actual target precision. For this reason, in the remainder of the paper we focus exclusively on the log-trained models as a basis for our algorithm selector.

% \begin{figure}
%     \centering
%     \includegraphics[width=0.48\textwidth,trim=160mm 5mm 40mm 5mm,clip]{figures/Log_Selection_Matrix.pdf}
%     \caption{Heatmap showing in how many (out of $1\,200$) runs each algorithm is selected to switch to by the logarithmic model, based on the amount of budget given to this second part of the search.}
%     \label{fig:selected_alg_log}
% \end{figure}

\begin{figure*}
    \centering
     \begin{subfigure}[b]{0.49\textwidth}
         \centering
         \includegraphics[width=\textwidth]{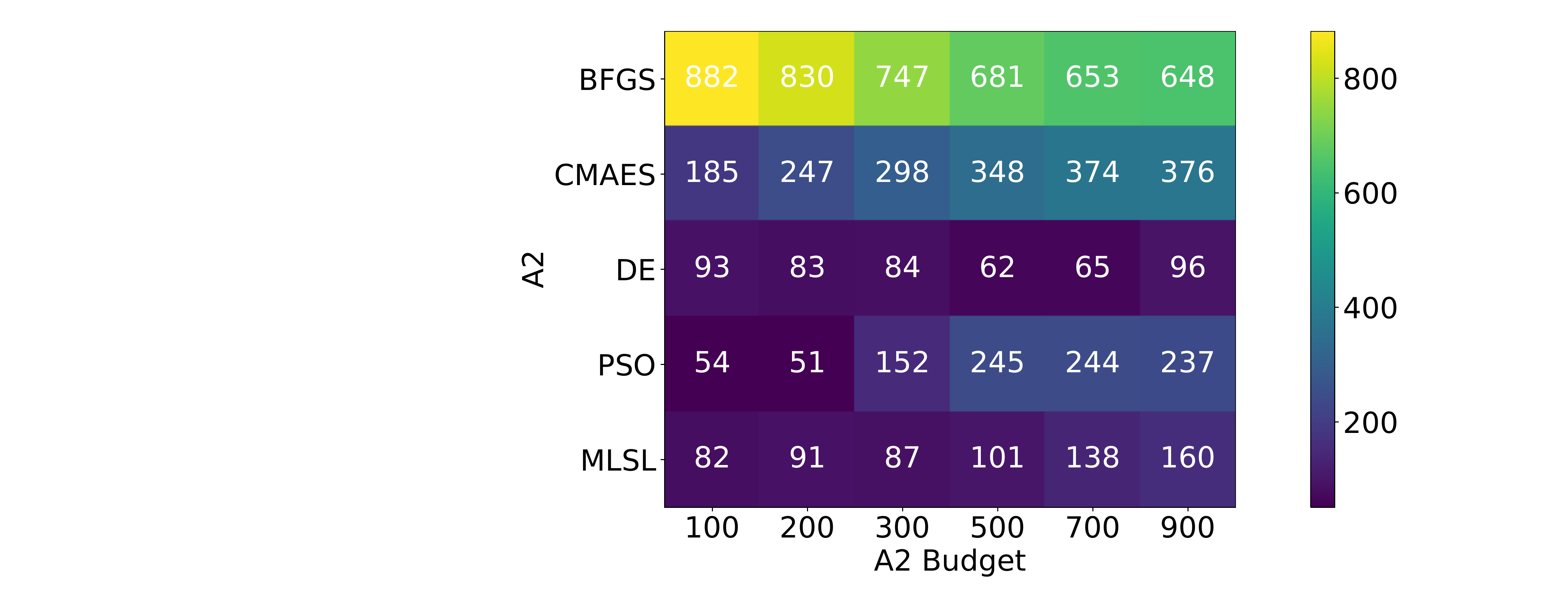}
        \caption{How often each solver is the best to switch to.}
        \label{fig:best_alg}
     \end{subfigure}
     \hfill
     \begin{subfigure}[b]{0.49\textwidth}
         \centering
         \includegraphics[width=\textwidth]{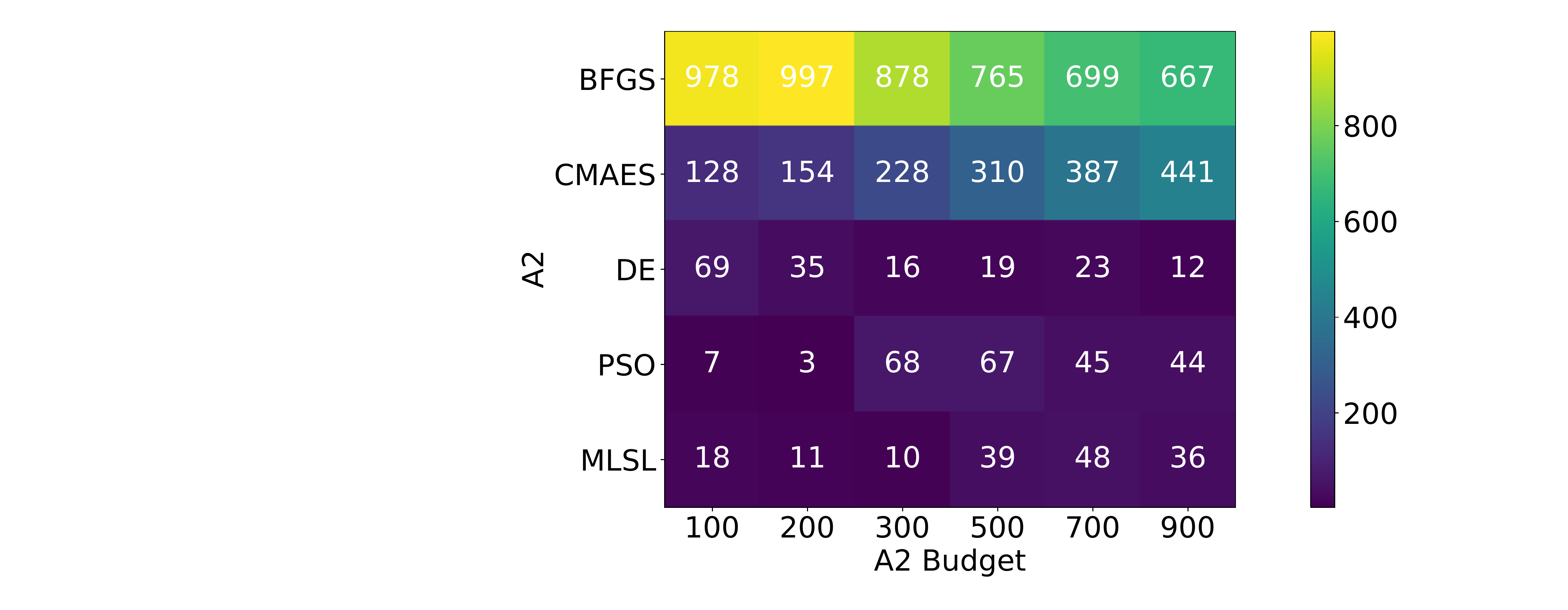}
        \caption{How often each solver is actually selected.}
        \label{fig:selected_alg_log}
     \end{subfigure}
     \caption{Heatmap showing in how many (out of $1\,200$) runs each algorithm is the best one to switch to (left) and is selected to switch to by the logarithmic model (right), based on the amount of budget given to this second part of the search. Results are capped at $10^{-8}$ target precision, which can lead to ties. The number of best algorithms per budget can therefore be larger than $1\,200$.}
    \label{fig:loss_log_model_with_bfgs}
\end{figure*}

\subsection{Evaluation of the Algorithm Selector}
\label{ssec:evaluation}

Once the predictions from all regression models are available, the next step is to select the best algorithm for each performed run on every problem instance. To this end, we choose the algorithm whose regression model provides the best predicted performance value for that run (i.e., we refer to this algorithm as the selected algorithm). For each run, we also identify the best algorithm based on the raw performance data (i.e., we refer to this algorithm as the best algorithm or the \emph{virtual best solver}).

To evaluate the performance of the algorithm selector, for each run individually, we compute the difference between the target precision of the selected algorithm $F_{\mathcal{A}}$ and that of the best one $F_{\mathcal{A}^{*}}$ (for that particular run). More precisely, we consider the difference after taking the logarithm of the achieved target precision:
$
\mathcal{L}(\mathcal{A}, \mathcal{A}^{*}) = \log(F_{\mathcal{A}}) -  \log(F_{\mathcal{A}^{*}}).
$
This gives us one performance measure per run, and we mainly investigate the distribution of these ``\emph{losses}'' over all $1\,200$ runs, which we compare to that of the five algorithms.

%%%%%%%%%%%%%%%%%%%%%
%%%%%%%% Results %%%%%%
%%%%%%%%%%%%%%%%%%%%%
\section{Results and Discussion}
\label{sec:results}

We first present the results of our trajectory-based algorithm selection approach for the full algorithm portfolio. % in Sec.~\ref{ssec:resultsfull}. 
Since BFGS clearly dominates several of the settings, we also consider what happens if we exclude it from the portfolio.  
% (Sec.~\ref{ssec:resultsnoBFGS}).

\begin{figure*}
    \centering
     \begin{subfigure}[b]{0.49\textwidth}
         \centering
         \includegraphics[trim={0 2cm 0 0},clip,width=\textwidth]{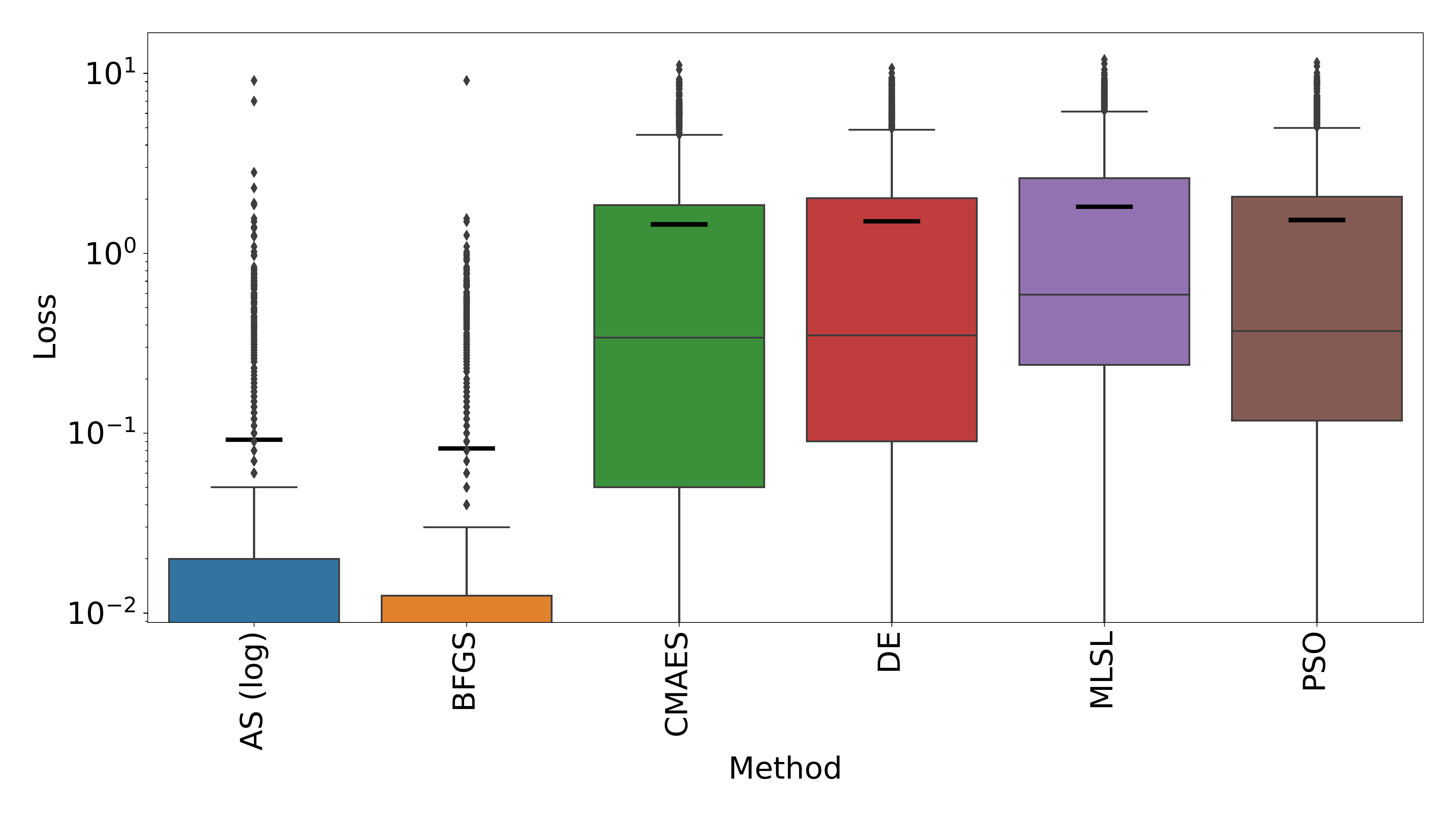}
         \caption{Budget 100}
         \label{fig:loss_B100}
     \end{subfigure}
     \hfill
     \begin{subfigure}[b]{0.49\textwidth}
         \centering
         \includegraphics[trim={0 2cm 0 0},clip,width=\textwidth]{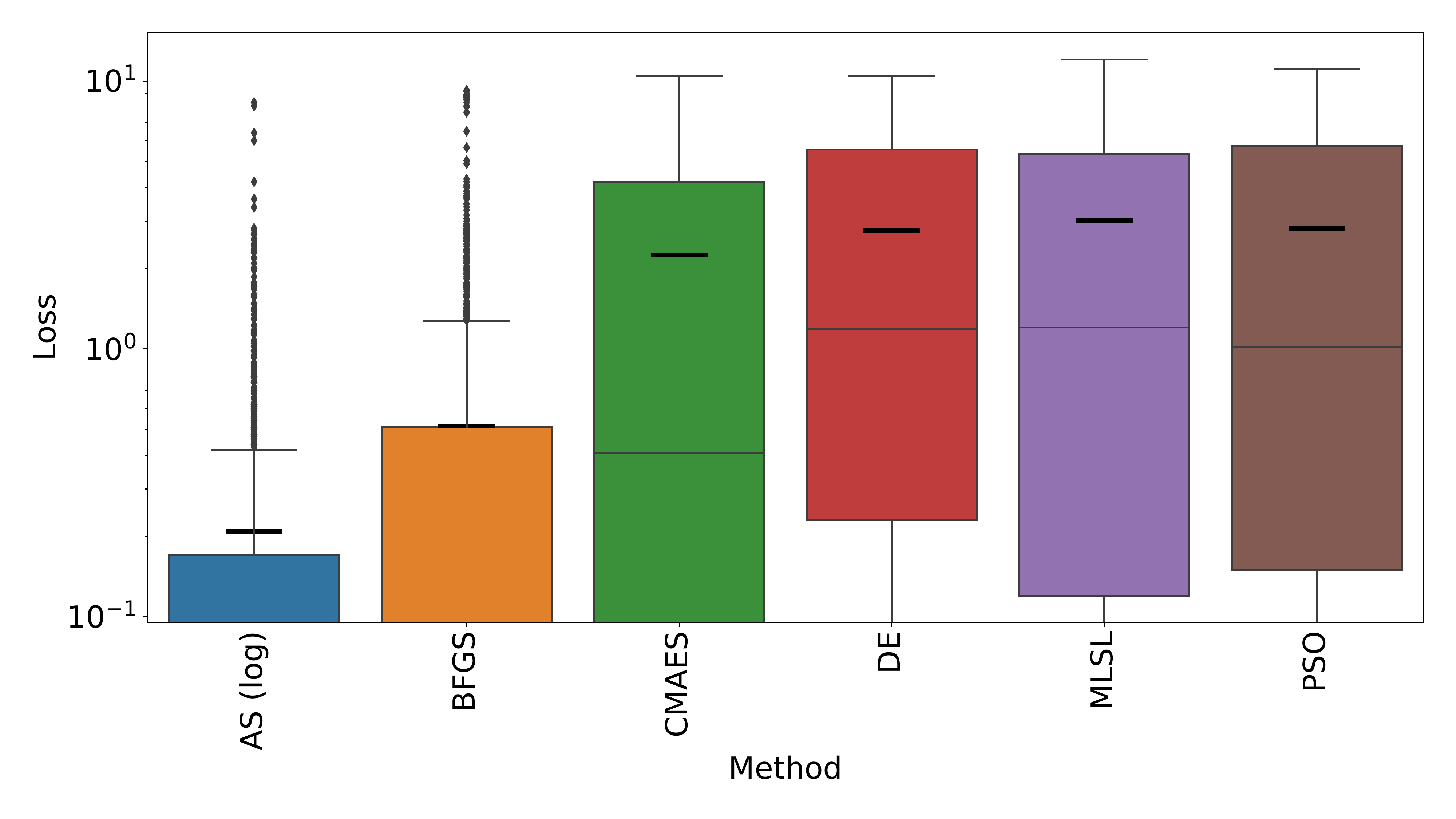}
         \caption{Budget 900}
         \label{fig:loss_B900}
     \end{subfigure}
     \caption{Loss (measured as difference between the achieved target precision and that of the virtual best solver, in log-performance space) of the logarithmic algorithm selection model and each of the five individual algorithms, for different budgets of the second part of the search. The thick black bar represents the mean loss for each method.}
    \label{fig:loss_log_model_with_bfgs}
\end{figure*}

\textbf{Full Portfolio.}
% \label{ssec:resultsfull}
Fig.~\ref{fig:loss_log_model_with_bfgs} shows the loss (computed as described in Sec.~\ref{ssec:evaluation}) of the five algorithms and our trajectory-based algorithm selector, for two \emph{A2} budgets, $100$ and $900$. As already visible in Fig.~\ref{fig:perf_curves}, the performance differences between the algorithms is not very pronounced for the small budget, with a vast majority of losses smaller than one order of magnitude. BFGS nevertheless clearly outperforms the other four algorithms. Our algorithm selector selects BFGS on 972 out of all $1\,200$ runs (see Fig.~\ref{fig:selected_alg_log}). It performs slightly worse than BFGS, i.e., we do not gain in this setting from the landscape-aware selection.

For budget 900 the situation is different. Here, BFGS is still the best solver when considering the loss distribution over all $1\,200$ runs. However, CMA-ES and MLSL are best for 252 and 167 runs, respectively (see Fig.~\ref{fig:best_alg}), and our algorithm selector manages to distinguish between these runs in at least some cases. To further probe into the decision of the algorithm selector, we present a confusion matrix in Tab.~\ref{tab:confusion900}. Our selector has chosen BFGS 667 times in total, and in 487 of these cases this choice was optimal. For 48 runs it would have been better to select CMA-ES, etc. 

\begin{table}[]
\centering
\caption{Confusion matrix for our algorithm selector for \emph{A2} budget 900. Total number is less than $1,200$ because we did not assign a confusion when more then one algorithm different from the selected one had (equal) best performance.\label{tab:confusion900}
}
\begin{tabular}{l|lllll|l}
          & \multicolumn{5}{|c}{\textbf{Selected algorithm}} &                           \\
\textbf{True (single) best} & BFGS & CMA-ES & DE & MLSL & PSO & \textbf{Total} \\
\hline
BFGS	& 	487	& 	44	& 	1	& 	1	& 	2	& 	\textbf{535}	\\
CMA-ES	& 	48	& 	252	& 	3	& 	4	& 	8	& 	\textbf{315}	\\
DE	& 	11	& 	18	& 	1	& 	3	& 	1	& 	\textbf{34}	\\
MLSL	& 	82	& 	66	& 	2	& 	33	& 	4	& 	\textbf{187}	\\
PSO	& 	33	& 	61	& 	3	& 	0	& 	20	& 	\textbf{117}	\\
\hline
\textbf{Total ($1\,188$)}	& 	\textbf{661}	& 	\textbf{441}	& 	\textbf{10}	& 	\textbf{41}	& 	\textbf{35}	& 	
\end{tabular}
\end{table}

\textbf{Excluding BFGS.} 
% \label{ssec:resultsnoBFGS}
We have seen above that the settings are largely dominated by BFGS, which is the best among the five algorithms for 648 (\emph{A2} budget $900$) up to 882 (\emph{A2} budget $100$) out of the $1\,200$ runs. We therefore analyze how the results change if we exclude BFGS from our portfolio. Note that we do not need to retrain our regression models for this setup, as they were trained for each algorithm individually.

% \begin{figure}
%     \centering
%     \includegraphics[width=0.48\textwidth,trim=120mm 5mm 40mm 5mm,clip]{figures/Best_Alg_Matrix_noBFGS.pdf}
%     \caption{Heatmap showing in how many (out of $1\,200$) runs each algorithm (excluding BFGS) is the best one to switch to, based on the amount of budget given to this second part of the search. Results are capped at $10^{-8}$ target precision, the number of best algorithms per budget can therefore be larger than $1\,200$.}
%     \label{fig:noBFGSbest}
% \end{figure}

% \begin{figure}
%     \centering
%     \includegraphics[width=0.48\textwidth,trim=120mm 5mm 40mm 5mm,clip]{figures/Log_Selection_Matrix_NoBFGS.pdf}
%     \caption{Heatmap showing in how many (out of $1\,200$) runs each algorithm (excluding BFGS) is selected to switch to by the logarithmic model, based on the amount of budget given to this second part of the search.}
%     \label{fig:selected_alg_log_noBFGS}
% \end{figure}

\begin{figure*}
    \centering
     \begin{subfigure}[b]{0.49\textwidth}
         \centering
         \includegraphics[width=\textwidth]{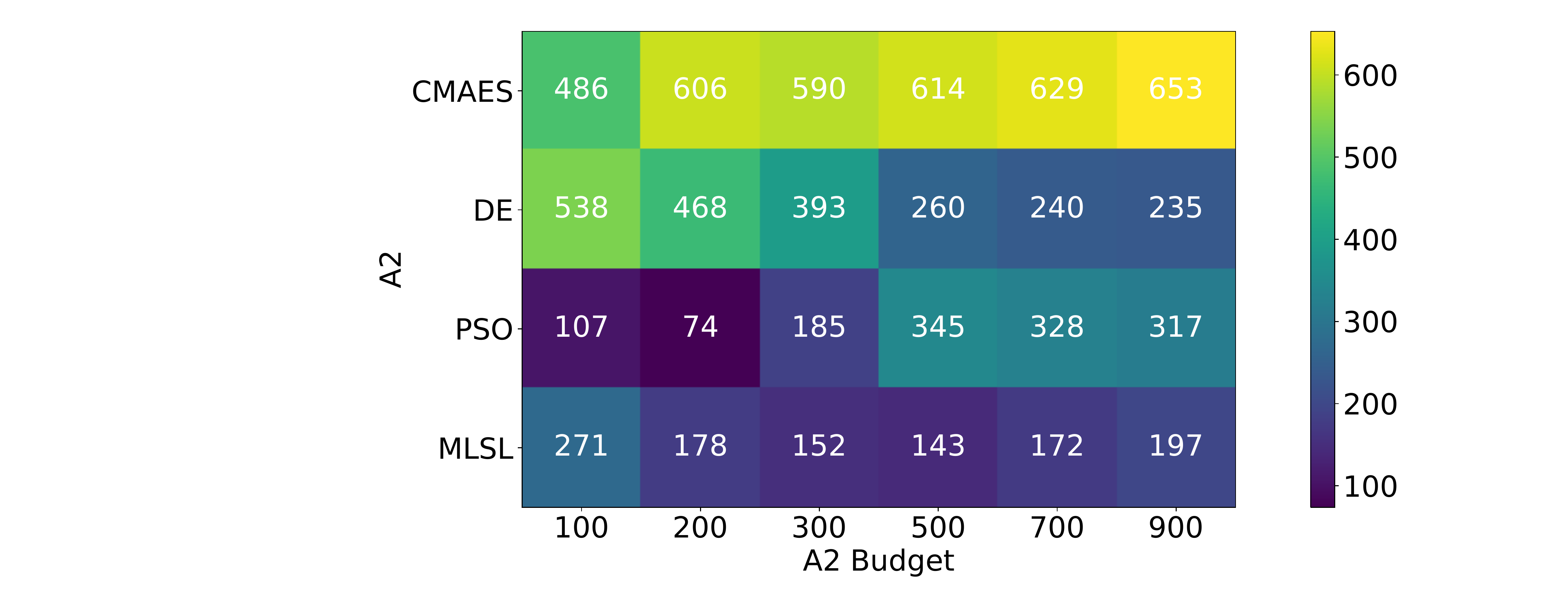}
        \caption{How often each solver, \emph{excluding BFGS}, is the best to switch to.}
        \label{fig:noBFGSbest}
     \end{subfigure}
     \hfill
     \begin{subfigure}[b]{0.49\textwidth}
         \centering
         \includegraphics[width=\textwidth]{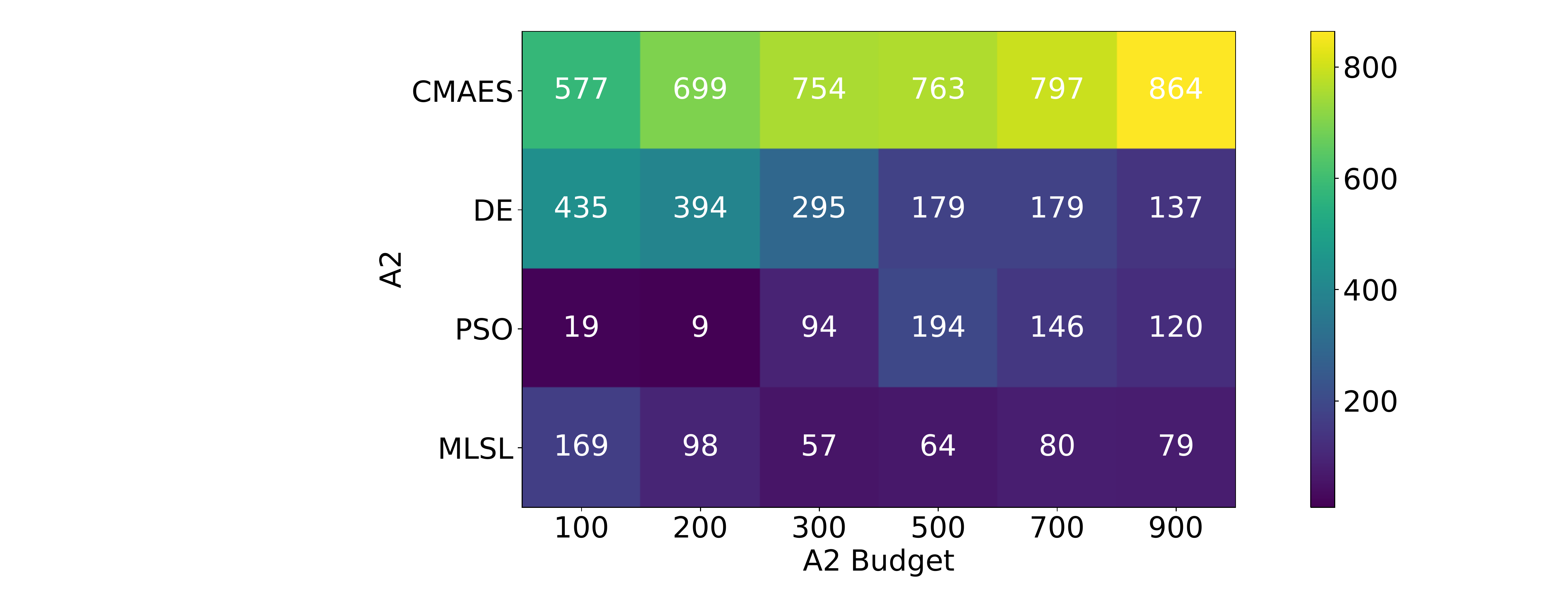}
        \caption{How often each solver, \emph{excluding BFGS}, is actually selected.}
        \label{fig:selected_alg_log_noBFGS}
     \end{subfigure}
     \caption{Heatmaps showing in how many (out of $1\,200$) runs each algorithm (excluding BFGS) is the best one to switch to (left) and is selected to switch to by the logarithmic model (right), based on the amount of budget given to this second part of the search. Results are capped at $10^{-8}$ target precision, the number of best algorithms per budget can therefore be larger than $1\,200$.}
    \label{fig:noBFGS}
\end{figure*}

\begin{figure*}
    \centering
     \begin{subfigure}[b]{0.49\textwidth}
         \centering
         \includegraphics[trim={0 2cm 0 0},clip, width=\textwidth]{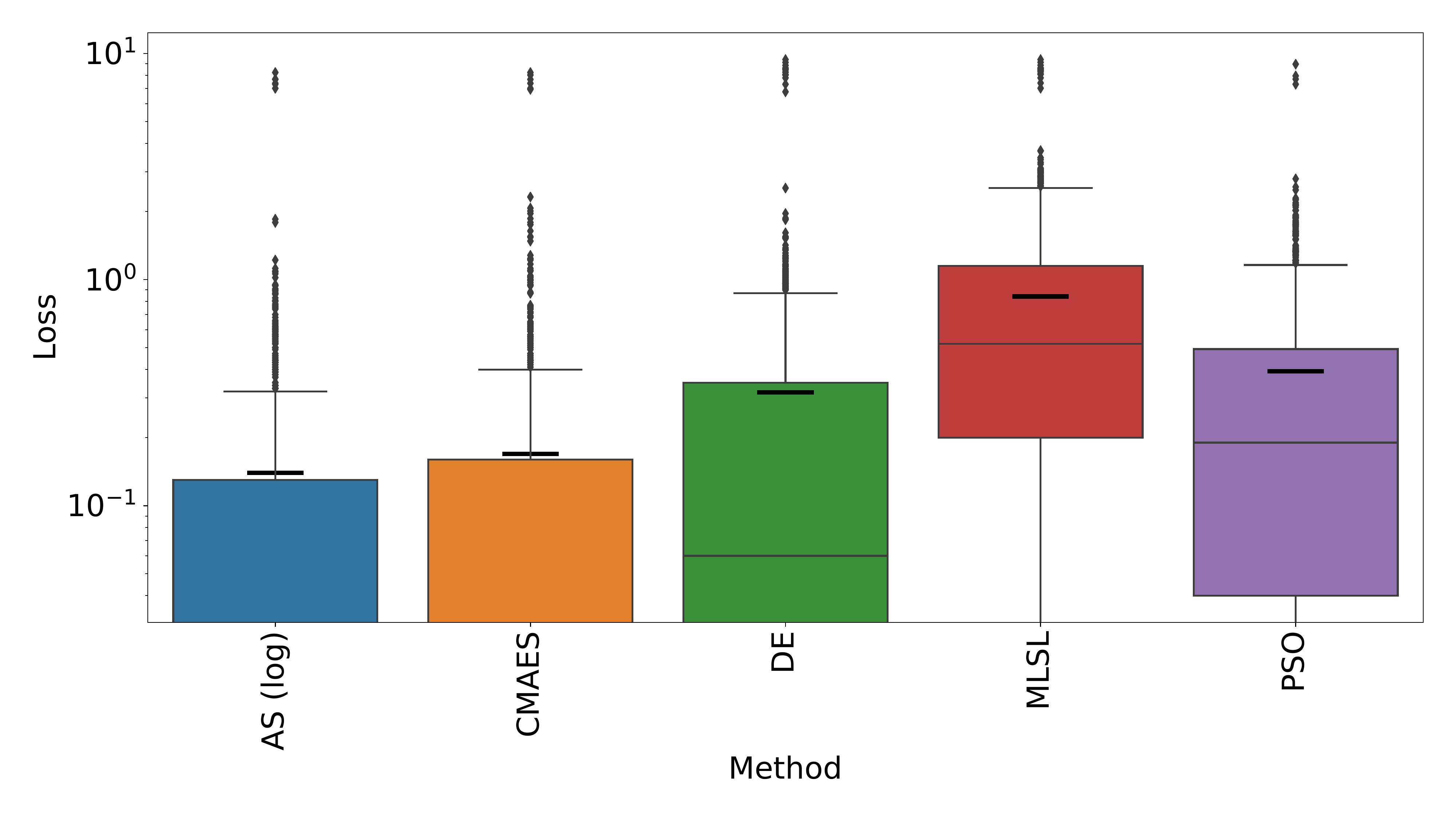}
         \caption{Budget 200}
         \label{fig:loss_B200_nobfgs}
     \end{subfigure}
     \hfill
     \begin{subfigure}[b]{0.49\textwidth}
         \centering
         \includegraphics[trim={0 2cm 0 0},clip,width=\textwidth]{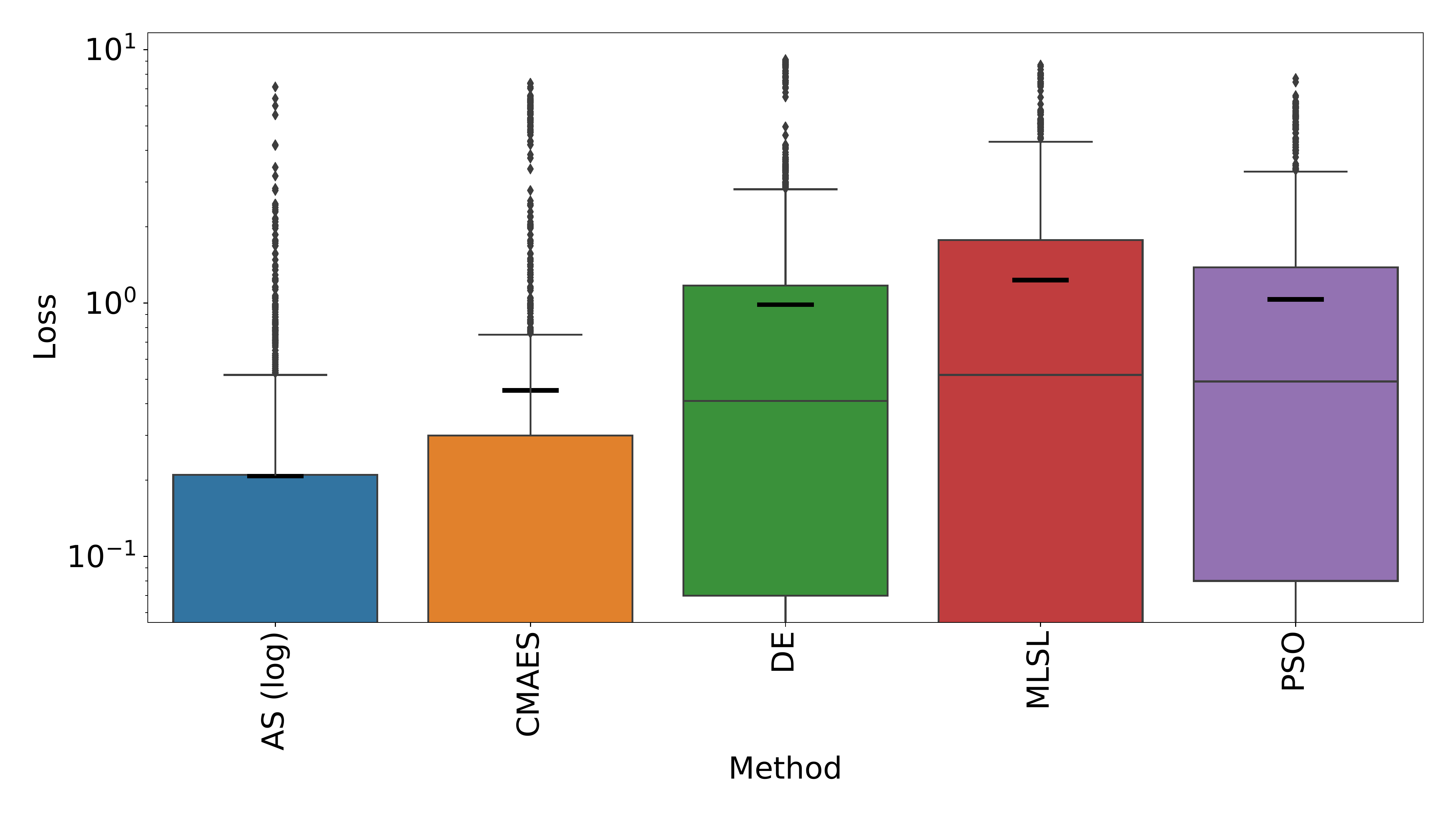}
         \caption{Budget 900}
         \label{fig:loss_B900_nobfgs}
     \end{subfigure}
     \caption{Loss (measured as difference between the achieved target precision and that of the virtual best solver, in log performance space) of the logarithmic algorithm selection model when excluding BFGS, and each of the 4 remaining individual algorithms, for different budgets of the second part of the search. The thick black bar represents the mean loss for each method.}
    \label{fig:boxplot-noBFGS}
\end{figure*}

Fig.~\ref{fig:noBFGSbest} summarizes how often each of the four algorithms is best (out of the same $1\,200$). We see that CMA-ES is now the dominating algorithm, however, to a much lesser extent as BFGS dominated the full portfolio. The algorithms selected by our algorithm selector seem to be equally balanced as the number of runs in which they are optimal. Note, though, that MLSL, DE, and PSO are selected much less often than the number of cases in which they are optimal suggests. That is, our algorithm selector often chooses CMA-ES. To evaluate the impact on the overall loss, we created again boxplots as in Fig.~\ref{fig:loss_log_model_with_bfgs}, for the same six \emph{A2} budgets as studied in the case with BFGS. Fig.~\ref{fig:boxplot-noBFGS} shows the results for \emph{A2} budgets 200 (left) and 900 (right). The results are very similar for all other \emph{A2} budgets: the loss of the CMA-ES is best among all four budgets, but the selector is better both in terms of mean performance (e.g., 0.14 vs. 0.17 for \emph{A2} budget 200 and 0.21 vs. 0.45 for \emph{A2} budget 900) and with respect to the 75\% percentile (0.13 vs. 0.16 for A2 budget 200 and 0.21 vs. 0.30 for \emph{A2} budget 900, respectively; the median is 0 for both the CMA-ES and the selector for most cases). 

% \begin{figure}[]
%     \centering
%     \includegraphics[width=.45\columnwidth]{prelim-figs/box-200.PNG}
%     \includegraphics[width=.45\columnwidth]{prelim-figs/box-900.PNG}
%     \caption{Distribution of loss values for the setting without BFGS and A2 budget $200$ (left) and $900$ (right).  
%     \carola{Figure to be replaced by Diederick}}
%     \label{fig:boxplot-noBFGS}
% \end{figure}

\section{Limitations of Our Approach}
\label{sec:limitations}

There are several limitations in our approach. First, we investigate only 10 runs per each algorithm on each problem instance and the results are computed on a per-run basis. In particular, an algorithm that happens to underperform in this particular run is assigned a large loss in our evaluation, even though its ``typical'' performance for the same setting may be much better than what the results of that one particular run suggests. This can of course also happen the other way around, i.e., an algorithm may appear to be much better than its ``typical'' performance.    
While we think that the overall large number of runs considered in our work helps to average out such unwanted outlier effects, a more robust experimental setup should be considered for future work.

In addition to this, we should note that the used warm-starting techniques are quite straightforward. While this is useful for dynamic algorithm selection in general, we could also extend the warm-starting procedure to better utilize the available information. This should lead to better overall performance of the switching algorithms. Furthermore, to this end, each of the considered algorithms require some level of warm-starting customization, which results in different warm-starting procedures being applied depending on the algorithm. Additional effort merits to be put towards defining a universal warm-starting procedure that can be employed independently of the algorithm's internal operating mechanism.

%%%%%%%%%%%%%%%%%%%%%
%%%%%%%% end %%%%%%
%%%%%%%%%%%%%%%%%%%%%
\section{Conclusions and Future Work}
\label{sec:conclusions}

We have shown that the trajectory-based selection is able to outperform all of the individual algorithms in this portfolio, given that there is sufficient complementarity in their performance.
Since our experimental pipeline makes use of a relatively small number of samples to determine the algorithm to switch to, without considering any algorithm-specific state features, it highlights the potential of the overall approach. 

Going forward, we will extend our work to settings in which a proper transfer of learned regression models needs to be performed. To this end, we will consider a transfer to the benchmark collections from the CEC competitions~\cite{cec2013,cec2014,cec2015,cec2017} and to the (artificial and real-world) problem suits available in \texttt{nevergrad}~\cite{nevergrad}. We also plan on extending our approach towards larger algorithm portfolios. Specifically, it would be good to focus on a portfolio which contains complementary algorithms, which show varying behavior on different problem instances. In addition, more research is required to define suitable ways to warm-start the algorithms with the information gathered by the first algorithm. 

As mentioned in Sec.~\ref{sec:setup}, we recorded several state variables of the CMA-ES, but we did not make use of them in this present study. We believe that the regression models can strongly benefit from this information; possibly not in the na\"ive way applied in~\cite{DBLP:conf/evoW/JankovicED21} (where only the final state variables at the time of the switch were used as features for the regression model), but by extracting information from the \emph{evolution} of the state variables during the first part of the optimization process, before the switch. Such an approach based on time-series analysis have been suggested in the literature~\cite{NobelWB21timeseries}. There, it was shown that features computed on evolution of the state variables of the CMA-ES can be used to accurately classify variants of the algorithm, and predict which of the BBOB problems was being optimized. Combining such an approach with the algorithm selection methodology presented in this work would be a promising direction of research. In addition, approaches with recurrent neural networks~\cite{uribarri2022dynamical} (i.e., long short-term memory) and transformers~\cite{woo2022etsformer} for predicting from longitudinal trajectory data should be considered to enrich the performance regression, which is a key component of our algorithm selection pipeline. 

Finally, an adaptive switching policy (as opposed to switching after a fixed number of evaluations as investigated in this present paper) is another important direction towards practical applicability and adoption of our trajectory-based landscape-aware algorithm selection approach.

%%%%%%%%%%%%%%%%%%%%%
%%%%%%%% refs %%%%%%
%%%%%%%%%%%%%%%%%%%%%
% \bibliographystyle{IEEEtran}
% \bibliography{references}

\end{document}